\documentclass[10pt,journal,compsoc]{IEEEtran}
\ifCLASSOPTIONcompsoc
  \usepackage[nocompress]{cite}
\else
  \usepackage{cite}
\fi

\ifCLASSINFOpdf
\else
\fi

\usepackage{epsfig}
\usepackage{graphicx}
\usepackage{amsmath}
\usepackage{amssymb}

\usepackage{algorithm}
\usepackage{algorithmic}
\usepackage{amsmath}

\usepackage{tabularx}
\usepackage{diagbox}
\usepackage{footnote}
\usepackage{threeparttable}

\usepackage{adjustbox}
\usepackage{textcomp}
\usepackage{array,multirow,graphicx}
\usepackage{makecell}
\usepackage{color}
\usepackage{hyperref}
\hypersetup{colorlinks,linkcolor={red},citecolor={green},urlcolor={red}}  
\usepackage{booktabs}

\def\eg{\emph{e.g.}} 
\def\ie{\emph{i.e.}}

\def\etal{\emph{et al.}}

\begin{document}
%
\title{AlignSeg: Feature-Aligned Segmentation Networks}
%
%
%
%

\author{Zilong Huang,
        Yunchao Wei,~\IEEEmembership{Member,~IEEE},
        Xinggang Wang,~\IEEEmembership{Member,~IEEE},
        Wenyu Liu,~\IEEEmembership{Senior Member,~IEEE},
        Thomas S. Huang,~\IEEEmembership{Life Fellow,~IEEE},
        and Humphrey Shi,~\IEEEmembership{Senior Member,~IEEE}
\IEEEcompsocitemizethanks{\IEEEcompsocthanksitem Z. Huang, X. Wang and W. Liu are with the School of Electronic Information and Communications, Huazhong University of Science and Technology, Wuhan 430074, China (e-mail: hzl@hust.edu.cn, xgwang@hust.edu.cn, liuwy@hust.edu.cn).
\IEEEcompsocthanksitem  Y. Wei is with AAII, University of Technology Sydney, Sydney, Australia (e-mail: wychao1987@gmail.com).
\IEEEcompsocthanksitem  T. S. Huang was with the Department of ECE and Beckman Institute, UIUC, Urbana, IL 61801 USA (e-mail: t-huang1@illinois.edu).
\IEEEcompsocthanksitem H. Shi is with Computer Science at the University of Oregon, ECE at UIUC and Picsart AI Research (e-mail: hshi10@illinois.edu).
}
\thanks{Corresponding author: Xinggang Wang}
}

%
%

\markboth{}%
{Shell \MakeLowercase{\textit{et al.}}: Bare Demo of IEEEtran.cls for Computer Society Journals}
%



\IEEEtitleabstractindextext{%
\begin{abstract}
Aggregating features in terms of different convolutional blocks or contextual embeddings has been proven to be an effective way to strengthen feature representations for semantic segmentation. However, most of the current popular network architectures tend to ignore the misalignment issues during the feature aggregation process caused by 1) step-by-step downsampling operations, and 2) indiscriminate contextual information fusion. In this paper, we explore the principles in addressing such feature misalignment issues and inventively propose Feature-Aligned Segmentation Networks (AlignSeg). AlignSeg consists of two primary modules, \ie, the Aligned Feature Aggregation (AlignFA) module and the Aligned Context Modeling (AlignCM) module. First, AlignFA adopts a simple learnable interpolation strategy to learn transformation offsets of pixels, which can effectively relieve the feature misalignment issue caused by multi-resolution feature aggregation. Second, with the contextual embeddings in hand, AlignCM enables each pixel to choose private custom contextual information adaptively, making the contextual embeddings be better aligned. We validate the effectiveness of our AlignSeg network with extensive experiments on Cityscapes and ADE20K, achieving new state-of-the-art mIoU scores of 82.6\% and 45.95\%, respectively. Our source code is available at \url{https://github.com/speedinghzl/AlignSeg}.
\end{abstract}

}

\maketitle

\IEEEdisplaynontitleabstractindextext

%
\IEEEpeerreviewmaketitle

\IEEEraisesectionheading{\section{Introduction}\label{sec:introduction}}

%
%
%
%

\IEEEPARstart{A}{s} one of the most fundamental tasks in computer vision, semantic segmentation provides significant impacts on various real-world applications, such as autonomous driving~\cite{fritsch2013new}, augmented reality~\cite{azuma1997survey}, and image editing~\cite{evening2012adobe}.
Recently, the development in deep learning \cite{krizhevsky2012imagenet} brings great leap-forwards to semantic segmentation with the fully convolutional architectures~\cite{chen2018deeplab,chen2018encoder,long2015fully,yu2016multi}.

We first review two common knowledge in semantic segmentation: 1) Deep features in different convolutional blocks are equipped with different characteristics, \ie, early features have rich spatial information yet lack semantic information due to their small receptive fields, and vice versa; 2) global context can provide useful information to alleviate unexpected confusion in predicting challenging regions. These two principles have been extensively explored to construct segmentation networks by either aggregating features across different convolutional blocks~\cite{chen2017rethinking,yu2018deep,ronneberger2015u} or introducing informative contextual features~\cite{chen2018deeplab, zhao2017pyramid,fu2019dual}. These explorations have made impressive contributions to semantic segmentation, however, one significant issue, \ie, \textbf{feature misalignment}, always fails to be considered.

We observe that feature misalignment issue mainly comes from two aspects: the step-by-step downsampling operations in almost all Convolutional Neural Networks (CNNs) and the indiscriminate treatment in contextual information fusion. To be specific, the downsampling operations will result in spatial misalignment issues when aggregating the late upsampled features with the early high-resolution ones as adopted by~\cite{chen2017rethinking,yu2018deep}. Besides, the common practice for introducing contextual information is to aggregate homogeneous contextual embeddings for all the pixels without considering their inherent difference, leading the unexpected misclassification caused by the falsely aligned contextual features.

\begin{figure}[!t]
    \centering
    \includegraphics[width=0.8\linewidth]{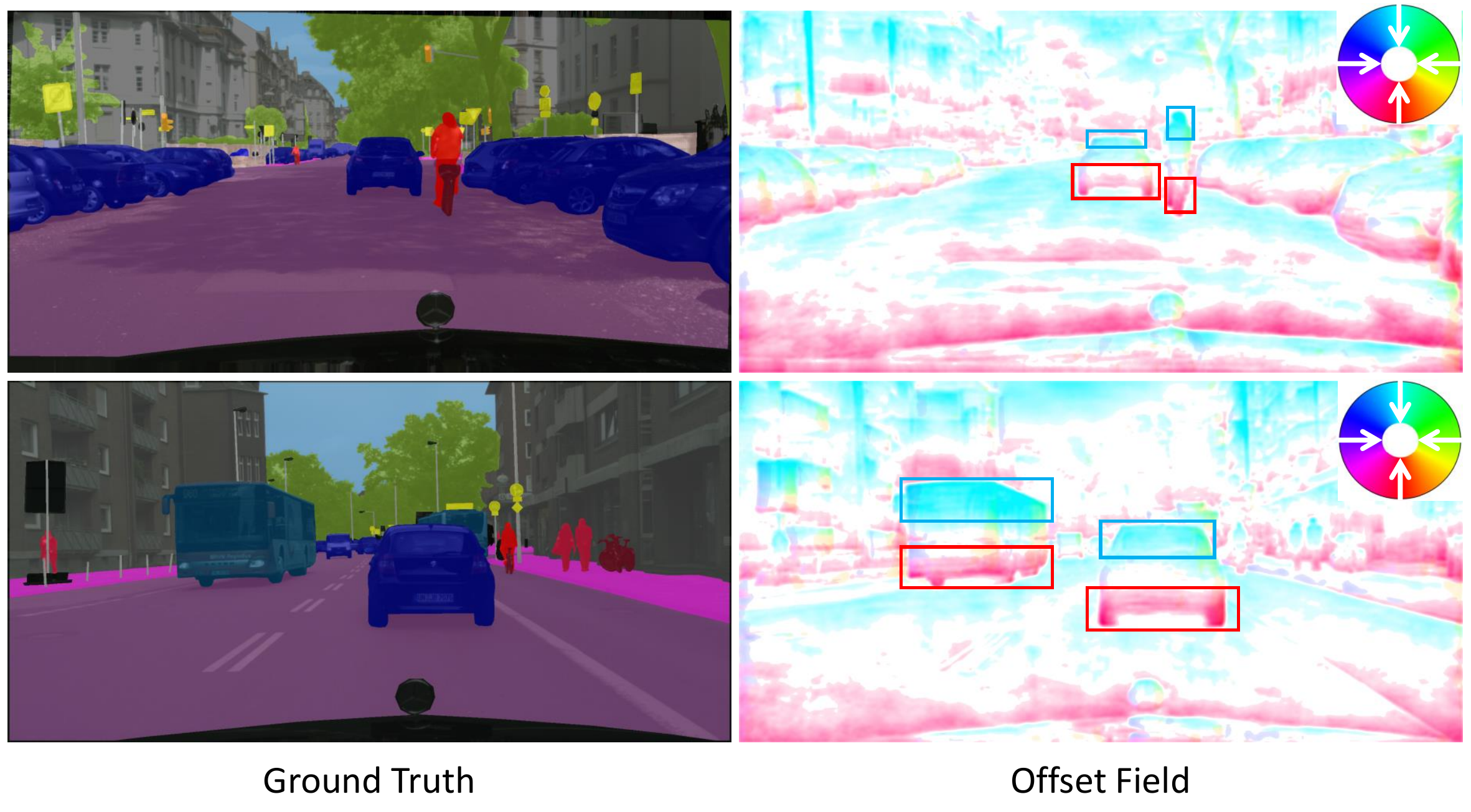}
    \caption{Visualizations of the learned 2D transformation offsets for feature alignment. Features of the regions with red/blue color inside the red/blue rectangles are aligned with features of their above/below regions.}
    \label{fig:offset}
    \vspace{-5mm}
\end{figure}

Accordingly, we propose Feature-Aligned Segmentation (AlignSeg) Networks. AlignSeg contains two primary modules: Aligned Feature Aggregation (AlignFA) and Aligned Context Modeling (AlignCM). AlignFA is used to precisely align both high-resolution and low-resolution feature maps for low-level and high-level information aggregation, \ie, multi-resolution feature aggregation. AlignCM performs adaptive, selective, and rich context pooling, which is in need of dense prediction tasks, \eg, semantic segmentation.


Specifically, to deal with misalignment caused by the downsampling operations, AlignFA is proposed for more precise multi-resolution feature aggregation. Here we choose the bottom-up feature aggregation architecture for its effectiveness~\cite{yu2018deep,pang2019towards}. On one hand, we employ two convolutional kernels to learn 2D transformation offsets for low-resolution features. Instead of the regular grid sampling of the bilateral interpolation in most current state-of-the-arts \cite{chen2018encoder,lin2017refinenet,ronneberger2015u}, these 2D transformation offsets will help to guide the feature alignment. Accordingly, the aligned features and the original high-resolution features are spatially aligned. Despite being a seemingly minor change, our learnable alignment approach provides more accurate predictions, especially for boundary regions. As shown in Fig.~\ref{fig:offset}, our approach learns to align the features in boundary regions with the features in the object regions, which helps to make correct predictions for these boundary regions.
On the other hand, we also learn 2D transformation offsets for high-resolution features. Because the high-resolution low-level feature may contain excessive details in textures or shape within the objects which could make it difficult for feature aggregation. These 2D transformation offsets will help to reduce excessive details within objects.


To collect rich contextual information, AlignCM utilizes a pooling based method to increase the ability to collect contextual information. But the pooling based methods, \eg, PSPNet~\cite{zhao2017pyramid}, aggregate homogeneous contextual information for all image pixels in a non-adaptive manner. To adaptively select the context for each pixel, we integrate a learnable alignment approach with pooling based context modeling methods. First of all, the local feature is fed into the pooling layer to generate the context feature with the spatial size of $K\times K$. We also employ two convolutional kernels to learn 2D transformation offsets over the context feature and the local feature. These 2D transformation offsets will help to guide the context feature to adjust spatial position in line with the local feature. Accordingly, the aligned feature and the local feature are context aligned. Finally, the local feature and aligned feature are concatenated as the final output feature. 

To demonstrate the effectiveness of our approach, we conduct elaborate experiments on the challenging Cityscapes~\cite{cordts2016cityscapes} and ADE20K~\cite{zhou2017scene} datasets. Our approach obtains competitive results, \eg, 82.6\% and 45.95\% mIoU scores on Cityscapes test set and ADE20K validation set, respectively. To summarize, our contributions include
\begin{itemize}
    \item We propose an Aligned Feature Aggregation module which uses a learnable alignment approach to align multi-resolution features during aggregation, which addresses the feature misalignment issue effectively.
    \item We introduce an Aligned Context Modeling module which enables each pixel to choose private custom contextual information.
    \item Our approach obtains 82.6\% and 45.95\% mIoU scores on Cityscapes test set and ADE20K validation set respectively.
\end{itemize}

The rest of this paper is organized as follows. We first
review related work in Section~\ref{Related work} and describe the architecture of our network in Section~\ref{CC network}. In Section~\ref{Experiments}, ablation studies are given and experimental results are analyzed. Section~\ref{Conclusion} presents our conclusion and future work.

\begin{figure*}[!t]
    \centering
    \includegraphics[width=0.9\linewidth]{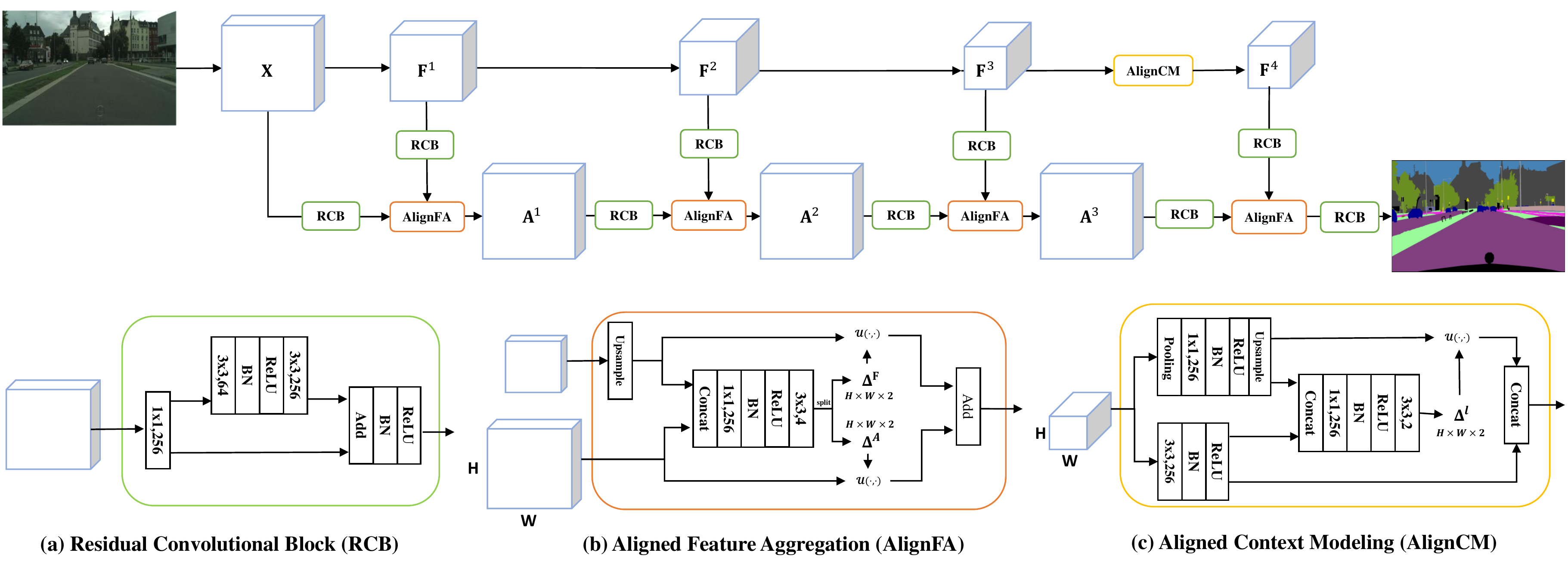}
    \caption{The architecture of our bottom-up aggregation with the Aligned Feature Aggregation module (b) and Aligned Context Modeling module (c). RCB indicates the common Residual Convolution Block (a). There are two bottom-up pathways in the architecture, and the feature alignment aggregation is used to aggregate features from the two pathways. For simplification, we omit the convolutional blocks in the top pathway. } 
    \label{fig:architecture} 
    \vspace{-6mm}
\end{figure*}

\vspace{-2mm}
\section{Related work} \label{Related work}

\noindent\textbf{Semantic segmentation.} Semantic segmentation has attracted great interests in recent years~\cite{badrinarayanan2017segnet,chen2018searching,chen2018deeplab,chen2017rethinking,chen2018encoder,lin2017refinenet,long2015fully,huang2019semantic,pohlen2017full,yu2018learning,yu2016multi,zhao2017pyramid,jiao2019geometry,chiu2020agriculture}. It is important to learn features that capture both rich spatial and semantic information for semantic segmentation. Recent progresses mainly follow two strategies for this problem,
\ie, feature aggregation~\cite{badrinarayanan2017segnet,chen2018encoder,lin2017refinenet,pohlen2017full,yu2018learning} and context modeling~\cite{chen2018deeplab,chen2017rethinking,yu2016multi}.

\noindent\textbf{Feature Aggregation.}
In feature aggregation, features from different network stages are aggregated to capture both rich spatial and semantic information. The most common used feature aggregation architecture is the encoder-decoder architecture which performs top-down aggregation, \eg, SegNet~\cite{badrinarayanan2017segnet}, U-Net~\cite{pohlen2017full}, RefineNet~\cite{lin2017refinenet}, ACNet~\cite{fu2019adaptive} and DeepLabv3+~\cite{chen2018encoder}. Another commonly used feature aggregation architecture is the bottom-up architecture, \ie, FRRN~\cite{pohlen2017full}, DLA~\cite{yu2018deep} and SeENet~\cite{pang2019towards}. The feature aggregation based approaches aggregate multi-resolution features from different convolutional blocks, which may have a potential issue, \ie, feature misalignment.

\noindent\textbf{Feature Alignment.} 
To solve the misalignment between network inputs and outputs caused by RoIPool~\cite{girshick2015fast},
He~\etal~\cite{he2017mask} proposes a RoIAlign approach to align output proposal features with input proposals. SegNet~\cite{badrinarayanan2017segnet} used pooling indices computed in the max-pooling step of the corresponding encoder to perform non-linear upsampling. Lu~\cite{lu2019indices} proposed the IndexNet to learn indices for guiding the pooling and upsampling operators. To enhance the ability of spatial invariance in CNN, Jaderberg \etal~\cite{jaderberg2015spatial} introduce a new learnable module to explicitly allow the spatial manipulation of data or feature. GUM~\cite{mazzini2018guided} proposed a guided upsampling module to learn 2D transformation offsets for each position. Our approach shares a similar aspect with the three approaches~\cite{jaderberg2015spatial, mazzini2018guided, dai2017deformable} in that we also learn 2D transformation offsets. But unlike these three approaches, our goal is to align multi-resolution features from different convolutional blocks for better aggregation, thus our approach requires to take features from different blocks as inputs rather than from a single block in~\cite{jaderberg2015spatial}. Besides, we design the Aligned Feature Aggregation module which aligns low resolution and high resolution features simultaneously.


\noindent\textbf{Context Modeling.}
For context modeling, the ASPP module was proposed in DeepLab~\cite{chen2018deeplab} by adopting parallel convolutional layers with different dilation,  PSPNet~\cite{zhao2017pyramid} was proposed to perform spatial pooling at several grid scales for context modeling. However, these methods utilize homogeneous contextual information for all image regions in a non-adaptive manner. Thus, some works~\cite{huang2019ccnet,fu2019dual,yuan2018ocnet,zhao2018psanet} adopt an attention mechanism to model the contextual information in a flexible and adaptive manner. Different from these attention based methods, we propose a new context modeling method that integrates the learnable alignment function into a pooling based method to select contextual information for each position in an adaptive manner.

{\noindent \textbf{AlignSeg \emph{vs.} DeformableNet~\cite{dai2017deformable} \& AlignDet~\cite{chen2019revisiting}} The three approaches all try to solve the misalignment issue in vision task, but they have three differences. \emph{Different motivations}: DeformableNet adaptively determines receptive field sizes for various objects. AlignDet aligns the features and its corresponding anchors. Our AlignSeg aims at aligning the multi-resolution features during aggregation. \emph{Different features in learning offset maps}: DeformableNet learns a offset map ($\mathbb{R}^{H \times W \times 18}$) from the feature to be aligned. ROIConv in AlignDet learns a anchor map ($\mathbb{R}^{H \times W \times 4}$) from the feature at previous layer. Our AlignFA learns an offset map ($\mathbb{R}^{H \times W \times 2}$) from the concatenated feature. \emph{Different align functions.} DeformableNet and AlignDet use deformable convolution and ROI Convolution respectively. They both contain learnable parameters. In contrast, our AlignSeg adopts a non-parameter align function.}

\vspace{-4mm}
\section{Approach} \label{CC network}

The overall network architecture with the feature alignment approach and context alignment approach is shown in Fig.~\ref{fig:architecture}.
As we can see, our network is a fully convolutional network~\cite{long2015fully}
which uses a bottom-up aggregation architecture, see Section~\ref{sec:buaa},
where the aggregation is performed by an Aligned Feature Aggregation module, see Section~\ref{sec:faa}. To enhance the local features with contextual information in an adaptive manner, we propose the Aligned Context Modeling module, see Section~\ref{sec:cam}.
Without loss of generality, we choose ResNets~\cite{he2016deep} as our backbone and other CNNs can also be chosen as the backbone in our approach.

\vspace{-2mm}
\subsection{Network Architecture}
\label{sec:buaa}

As shown in Fig.~\ref{fig:architecture},
taking an image $\mathbf{I} \in \mathbb{R}^{3 \times H \times W}$ as input,
where $3, H, W$ indicate the RGB channels, height, width of $\mathbf{I}$ respectively,
our bottom-up architecture first passes the image through some convolutional blocks\footnote{We say that the layers generating features of the same resolution belong to the same convolutional block.} to generate convolutional features $\mathbf{X}$.
Here we choose the outputs from the last layer of the res2 block as $\mathbf{X}$,
and thus $\mathbf{X} \in \mathbb{R}^{256 \times \frac{H}{4} \times \frac{W}{4}}$.
Then $\mathbf{X}$ is branched into two bottom-up pathways.
The two pathways produce a series of features $\{\mathbf{F}^{1}, ..., \mathbf{F}^{T}\}$ and $\{\mathbf{A}^{1}, ..., \mathbf{A}^{T}\}$ respectively,
where $T$ indicates the number of convolutional blocks for each pathway.
After that, the features $\mathbf{F}^{T}, \mathbf{A}^{T}$ from the final convolutional blocks of the two pathways are fused together to form the final features $\mathbf{F}$.
Finally, $\mathbf{F}$ is used to predict the pixel-wise segmentation results. The two pathways begin from the outputs from the last layer of the res2 block instead of other blocks, because this choice provides a good trade-off between much spatial information and less computation/memory cost. The details of the two pathways are described below.

The first bottom-up pathway is the same as the original CNN architecture, see the top pathway in Fig.~\ref{fig:architecture}.
More precisely, the features from later convolutional blocks are of reduced spatial resolutions and enriched semantic information.
For ResNets, there are three convolutional blocks after the res2 block. To further gather rich contextual information, we add an Aligned Context Modeling module as the fourth block. This module takes the output of res5 block as input. We choose outputs from the last layer of each block as $\{\mathbf{F}^{1}, \mathbf{F}^{2}, \mathbf{F}^{3}\, \mathbf{F}^{4}\}$, where $\mathbf{F}^{1} \in \mathbb{R}^{512 \times \frac{H}{8} \times \frac{W}{8}}, \mathbf{F}^{2} \in \mathbb{R}^{1024 \times \frac{H}{16} \times \frac{W}{16}}, \mathbf{F}^{3} \in \mathbb{R}^{2048 \times \frac{H}{32} \times \frac{W}{32}}, \mathbf{F}^{4} \in \mathbb{R}^{2304 \times \frac{H}{32} \times \frac{W}{32}}$.
Here we suppose $H, W$ can be divisible by 32 because we can pad images to ensure that $H, W$ are divisible by 32.

The second bottom-up pathway maintains spatial information in $\mathbf{X}$,
\ie, there is no downsampling operation in the second pathway, see the bottom pathway in Fig.~\ref{fig:architecture}.
More specifically,
the second bottom-up pathway produces a series of 
features $\{\mathbf{A}^{1}, \mathbf{A}^{2}, \mathbf{A}^{3}, \mathbf{A}^{4}\}$. The resolution and number of channels of $\{\mathbf{A}^{1}, \mathbf{A}^{2}, \mathbf{A}^{3}, \mathbf{A}^{4}\}$  are simply set to the same as $\mathbf{X}$.
Apart from taking features from $\mathbf{A}^{t-1}$,
features $\mathbf{F}^{t}$ from the first bottom-up pathway are also incorporated to obtain $\mathbf{A}^{t}, t \in \{1, 2, 3, 4\}$,
where $\mathbf{A}^{0} = \mathbf{X}$.
More specifically, following~\cite{yu2018learning,yu2018deep}, we first pass $\mathbf{A}^{t - 1}$ and $\mathbf{F}^{t}$ through a Residual Convolution Block to produce features $\tilde{\mathbf{A}}^{t - 1} \in \mathbb{R}^{256 \times \frac{H}{4} \times \frac{W}{4}}$ and $\tilde{\mathbf{F}}^{t} \in \mathbb{R}^{256 \times \frac{H}{2^{t+2}} \times \frac{W}{2^{t+2}}}$, respectively.
Then an Aligned Feature Aggregation module is used to aggregate features $\tilde{\mathbf{A}}^{t - 1}$ and $\tilde{\mathbf{F}}^{t}$, see Section~\ref{sec:faa}.
Therefore, semantic information in the features from the second bottom-up pathway is gradually enriched when $t$ is increased.
Finally, the feature is passed through several convolutional layers to produce the final segmentation map.

\vspace{0.1cm}
\noindent\textbf{Residual Convolutional Block.}
As shown in Fig.~\ref{fig:architecture}~(a),
in the residual convolutional block,
a $1 \times 1$ convolutional layer is first used to reduce channel dimensions to 256.
Then two $3 \times 3$ convolutional layers are used to enrich semantic information
and the residual connection~\cite{he2016deep} is used to alleviate the gradient vanishing/exploding problem.
Especially, the first $3 \times 3$ convolutional layer reduces channel dimensions to 64 to reduce computation.

\vspace{-2mm}
\subsection{Aligned Feature Aggregation}
\label{sec:faa}

As stated in Section~\ref{sec:buaa},
features from the first and second bottom-up pathways are aggregated
by an aligned feature aggregation function 
to enrich semantic information for the second pathway.
Here we introduce the function in detail.

The most straightforward approach for aggregation is to follow the previous approaches~~\cite{badrinarayanan2017segnet,chen2018encoder,long2015fully,pohlen2017full,ronneberger2015u,yu2018deep},
\ie, first upsampling $\tilde{\mathbf{F}}^{t}$ by the standard regular grid sampling based bilateral interpolation, and then concatenating or adding $\tilde{\mathbf{A}}^{t - 1}$ and the upsampled $\tilde{\mathbf{F}}^{t}$ together.
However, such an approach will result in spatial misalignment between $\tilde{\mathbf{A}}^{t - 1}$ and the upsampled $\tilde{\mathbf{F}}^{t}$,
which harms segmentation performance, especially for boundary regions.
In addition, the high-resolution low-level feature may contain the excessive details in textures or shapes within the objects which may make it difficult for feature aggregation.
To address these issues, we propose a learnable alignment approach to align the both upsampled $\tilde{\mathbf{F}}^{t}$ and $\tilde{\mathbf{A}}^{t - 1}$.


Here we first upsample $\tilde{\mathbf{F}}^{t}$ by the
standard regular grid sampling based bilateral interpolation.
Then the upsampled $\tilde{\mathbf{F}}^{t}$ and $\tilde{\mathbf{A}}^{t - 1}$ are concatenated together. After that the concatenated features are passed through several convolutional layers to predict the two offset maps $\mathbf{\Delta}^{F} \in \mathbb{R}^{2 \times \frac{H}{4} \times \frac{W}{4}}$ and $\mathbf{\Delta}^{A} \in \mathbb{R}^{2 \times \frac{H}{4} \times \frac{W}{4}}$. The offset maps $\mathbf{\Delta}^{F}$ and $\mathbf{\Delta}^{A}$ will be used to align  the low-resolution feature $\tilde{\mathbf{F}}^{t}$ and high-resolution feature $\tilde{\mathbf{A}}^{t - 1}$, respectively.
See Fig.~\ref{fig:architecture}~(b).
We use both $\tilde{\mathbf{A}}^{t - 1}$ and $\tilde{\mathbf{F}}^{t}$ to learn $\mathbf{\Delta}$,
because $\mathbf{\Delta}$ is used to align $\tilde{\mathbf{F}}^{t}$ with $\tilde{\mathbf{A}}^{t - 1}$ and the alignment cannot be accomplished by a single feature.

After obtaining offset maps, our feature alignment aggregation could be performed as follows,
\vspace{-1mm}
\begin{equation}
    \label{equ:aggregation}
    \begin{aligned}
        \mathbf{A}^{t} = \mathcal{U}(\operatorname{upsample}(\tilde{\mathbf{F}}^{t}), \mathbf{\Delta}^{F}) + \mathcal{U}(\tilde{\mathbf{A}}^{t - 1}, \mathbf{\Delta}^{A})\text{,}
    \end{aligned}
\end{equation}
where $\operatorname{upsample}$ denotes bilateral interpolation function, $\mathcal{U}(\cdot, \cdot)$ is the alignment function. Suppose the spatial coordinates of each position on the feature map $\mathbf{F}$ to be aligned are $\{(1, 1), (1, 2), ..., (H, W)\}$ and the offset map is $\mathbf{\Delta} \in \mathbb{R}^{2 \times H \times W}$. The $\mathbf{U}_{hw}$ is the output of the alignment function $\mathcal{U}(\mathbf{F}, \mathbf{\Delta})$ and alignment function is defined as follows:
\vspace{-2mm}
\begin{equation}
    \label{equ:interpolation}
    \begin{aligned}
        \mathbf{U}_{hw} = \mathop{\sum} \limits_{h^{\prime} = 1}^{H} \mathop{\sum} \limits_{w^{\prime} = 1}^{W} \mathbf{F}_{h^{\prime}w^{\prime}} & \cdot \max(0, 1 - | h + \Delta_{1hw} - h^{\prime} |) \\
        & \cdot \max(0, 1 - | w + \Delta_{2hw} - w^{\prime} |)\text{,}
    \end{aligned}
\end{equation}
which samples feature on position $(h + \Delta_{1hw}, w + \Delta_{2hw})$ of $\mathbf{F}$,
using the bilinear interpolation kernel,
where $\Delta_{1hw}, \Delta_{2hw}$ indicate the learned 2D transformation offsets for position $(h, w)$.
Notice that if we do not learn $\mathbf{\Delta}$ and set $\Delta_{1hw}, \Delta_{2hw}$ in Eq.~\eqref{equ:interpolation} to 0,
the alignment function could not modify the input feature $\mathbf{F}$ and the $\mathbf{U}$ is identical to $\mathbf{F}$.

\vspace{-2mm}
\subsection{Aligned Context Modeling} \label{sec:cam}
Contextual information can provide rich semantic guidance for overall scene images, thus rectifying misclassification
and inconsistent parsing results. The pooling based methods~\cite{zhao2017pyramid, liu2015parsenet} are widely used and effective for aggregating contextual information.
However, the pooling based
methods aggregate contextual information in a non-adaptive manner and the homogeneous contextual information is adopted by all image pixels.
Then we can exploit it to
adaptively aggregate contextual information. To this end, we propose
an Aligned Context Modeling (AlignCM) module which integrates a learnable alignment operator with a pooling based approach. 

See Fig.~\ref{fig:architecture}~(c), given an input feature map $\mathbf{F} \in \mathbb{R}^{C \times H \times W}$,  we use an average pooling following a convolution layer (with BN and ReLU) to generate a contextual feature $\mathbf{F^c} \in \mathbb{R}^{C \times K \times K}$, where $K$ is the bin size of the contextual feature. Meanwhile, we use a convolution layer followed by a batch norm layer and an activation layer to generate a local feature $\mathbf{F^l} \in \mathbb{R}^{C \times H \times W}$, which has the same resolution as $\mathbf{F}$.
The contextual feature $\mathbf{F^c}$ is upsampled by the standard regular grid sampling based bilateral interpolation. Then upsampled $\mathbf{F^c}$ and the local feature $\mathbf{F^l}$,
are concatenated together and passed through several convolutional layers to predict the two offsets $\mathbf{\Delta}^{t} \in \mathbb{R}^{2 \times H \times W}$ for each position. The offsets and the contextual feature are passed into the alignment function $\mathcal{U}(\cdot, \cdot)$ to align the contextual feature for each position. The details of the alignment function $\mathcal{U}(\cdot, \cdot)$ are introduced in Section~\ref{sec:faa}. Finally, the aligned contextual feature and the local feature are concatenated in channel dimension as the output, as shown in the Eq.~\ref{equ:cam}.
\begin{equation}
    \label{equ:cam}
    \begin{aligned}
        \operatorname{output} = \operatorname{concat}(\mathbf{F^l}, \mathcal{U}(\mathbf{F^c}, \mathbf{\Delta}^{t}))\text{.}
    \end{aligned}
\end{equation}

Both AlignFA and AlignCM use the learnable alignment function, AlignFA is proposed to align features of different stages, which has two inputs, a low-resolution high-level feature and a high-resolution low-level feature. AlignCM is proposed to augment feature by aggregating contextual information in an adaptive manner, the learnable alignment function is used to select the context obtained by the pooling operator for each pixel. 

\vspace{-3mm}
\section{Experiments} \label{Experiments}

We conduct elaborate
experiments on the Cityscapes dataset~\cite{cordts2016cityscapes} and the ADE20K dataset~\cite{zhou2017scene}.
Experimental results demonstrate that the proposed approach obtains the state-of-the-art performance on Cityscapes and ADE20K.
We finally report our results on COCO~\cite{lin2014microsoft}.

\vspace{-3mm}
\subsection{Datasets and Evaluation Metrics}

\noindent\textbf{Cityscapes} is a dataset for urban segmentation.
There are 5000 images with high-quality pixel-level annotations and 20000 images with coarse annotations in the dataset.
Each image is of $1024 \times 2048$ resolution.
Each pixel belongs to one of 19 classes.
In our experiments, only the 5000 images with high-quality pixel-level annotations are used,
where 2975, 500, and 1525 images are chosen for training, validation, and testing, respectively.

\noindent\textbf{ADE20K} is a recent scene parsing dataset.
Images in the dataset are densely labeled as 150 stuff/object classes.
The dataset is divided into 20K, 2K, and 3K images for training, validation, and testing, respectively.

\noindent\textbf{COCO} is a very challenging dataset for instance segmentation that contains 115k images over 80 categories for training, 5k images for validation and 20k images for testing.

\vspace{-3mm}
\subsection{Implementation Details}

\noindent\textbf{Network Structure.} 
Our approach utilizes ImageNet pre-trained ResNet-101~\cite{he2016deep} as the backbone.
The last fully-connected layer is removed and feature aggregation is started from the output of the \emph{res2} block. The standard BatchNorm~\cite{ioffe2015batch} layer is replaced by the Synchronize BatchNorm~\cite{rota2018place}
to collect the mean and standard-deviation of BatchNorm across multiple GPUs during training. \\

\noindent\textbf{Training Settings.}
Following the prior work~\cite{chen2018deeplab}, we employ a poly learning rate policy where the initial learning rate is multiplied by $ 1 - (\frac{iter}{max\_iter})^{power} $ with $power$ = 0.9.
The initial learning rate is set to 1e-2 for Cityscapes and 2e-2 for ADE20K.
The mini-batch size is set to 16 and 32 for Cityscapes and ADE20K respectively during stochastic gradient descent training. 
We use the momentum of 0.9 and a weight decay of 0.0001.
There are some differences in data augmentation for Cityscapes and ADE20K. The training images are augmented by first randomly scaling and then randomly cropping out the fixed size patches from the resulting images.
In addition, we also apply random horizontally flipping for data augmentation.


\vspace{-3mm}
\subsection{Experiments on Cityscapes}

\subsubsection{Ablation Study}
\label{sec:ablation_exp}
We first conduct ablation experiments to discuss the influence of different components,
including the proposed AlignFA and AlignCM.
We then discuss the speed, memory, and accuracy of different approaches.
We also provide the results by adding standard bells and whistles.
Without loss of generality, all results are obtained by training on the training set and evaluate on the validation (val) set.

As shown in Table~\ref{tab:cityscape_modules}, we use the bottom-up architecture with ResNet-101 as our baseline, which achieves 77.1\% mIoU on Cityscapes val set. Adding the Aligned Context Modeling module, donated as ``Baseline+AlignCM'', brings 2.3\% mIoU improvements. Adding Aligned Feature Aggregation module and Aligned Context Modeling module, donates as ``Baseline+AlignCM+AlignFA'' could further bring 1.1\% mIoU gain. The improvements have proven the effectiveness of the proposed modules.

    \begin{table}[!t]
    \renewcommand{\arraystretch}{1.3}
    \setlength{\tabcolsep}{1em}
    \caption{Ablation study for the proposed modules on the Cityscapes \emph{val}.}
        \label{tab:cityscape_modules}
        \centering \small
        \vspace{-3mm}
    \begin{tabular}{c c c c}
        \toprule[0.2em]
         AlignCM & AlignFA & mIoU (\%)  \\
        \toprule[0.2em]
                  &           & 77.1 \\
        $\surd$   &           & 79.4 \\
                  &$\surd$    & 78.6 \\
        $\surd$   &  $\surd$  & 80.5 \\
        \bottomrule[0.1em]
    \end{tabular}
    \vspace{-3mm}
    \end{table}
    
    \begin{table}[!t]
        \renewcommand{\arraystretch}{1.3}
        \setlength{\tabcolsep}{1.3em}
        \caption{Results for different feature aggregation methods on the Cityscapes \emph{val}.}
        \label{tab:cityscape_faa}
        \vspace{-3mm}
        \centering \small
        \begin{tabular}{lccc}
            \toprule[0.2em]
            Aggregation Methods & RGS & Deconv & AlignFA  \\
            \toprule[0.2em]
            mIOU (\%) & 79.4 & 79.7 & 80.5 \\
            \bottomrule[0.1em]
        \end{tabular}
        \vspace{-3mm}
    \end{table}
    
    \begin{table}[!t]
        \renewcommand{\arraystretch}{1.3}
        \setlength{\tabcolsep}{1.7em}
        \caption{Results for different align directions of AlignFA on the Cityscapes \emph{val}.}
        \label{tab:cityscape_faa_direction}
        \vspace{-3mm}
        \centering \small
        \begin{tabular}{ccc}
            \toprule[0.2em]
            low $\xrightarrow{align}{}$ high & high $\xrightarrow{align}{}$ low  & mIOU (\%)  \\
            \toprule[0.2em]
                       &           & 79.4\\
             $\surd$   &           & 79.7\\
             $\surd$   & $\surd$   & 80.5\\
            \bottomrule[0.1em]
        \end{tabular}
        \vspace{-4mm}
    \end{table}


\noindent\textbf{Aligned Feature Aggregation.}
We further study the influence of our Aligned Feature Aggregation module.
Results from different feature aggregation approaches are shown in Table~\ref{tab:cityscape_faa},
where RGS indicates using the standard regular grid sampling based interpolation before aggregation,
Deconv indicates using the Deconvolution~\cite{noh2015learning} based interpolation before aggregation,
and AlignFA indicates using our Aligned Feature Aggregation module described in Section~\ref{sec:faa}.
As we can see, our Aligned Feature Aggregation module obtains much better performance than others.
This is because our approach learns interpolation functions to align upsampled features and the original high-resolution features, which provides more accurate predictions for object boundaries. 
Meanwhile, we study the influence of the align directions. Here, we use ``$low\xrightarrow{align}{} high$'' to denote aligning low-level feature with the guide of the high-level feature. Aligning high-level feature
with the guide of the low-level feature is denoted as ``$high\xrightarrow{align}{} low$''. As shown in Table~\ref{tab:cityscape_faa_direction}, ``$low\xrightarrow{align}{} high$'' improves the performance by 0.3\% mIoU and adding ``$high\xrightarrow{align}{} low$'' could further bring 0.8\% mIoU gains.

\begin{table}[!t]
        \renewcommand{\arraystretch}{1.3}
        \setlength{\tabcolsep}{0.85em}
        \caption{Ablation study of w/ and w/o alignment for the context alignment module on the Cityscapes \emph{val} set.}
        \label{tab:cityscape_cam}
        \vspace{-3mm}
        \centering \small
        \begin{tabular}{lccc}
            \toprule[0.2em]
            Bin Size & 2 & 3 & 6  \\
            \toprule[0.2em]
             w/o Alignment   & 79.1 & 79.0 & 78.5\\
             w/ Alignment  & 79.3 & 79.4 & 79.0\\
            \bottomrule[0.1em]
        \end{tabular}
        \vspace{-4mm}
    \end{table}
    
  \begin{table}[!t]
    \renewcommand{\arraystretch}{1.3}
    \setlength{\tabcolsep}{0.15em}
    \centering
    \caption{Comparison of context modeling methods on the Cityscape \emph{val} set.}
    \label{tab:ablation_contextual_module}
    \vspace{-3mm}
    \centering \small
    \begin{tabular}{c c c c c c}
        \toprule[0.2em]
         Modules & Baseline & +PSP~\cite{zhao2017pyramid} & +ASPP~\cite{chen2018deeplab} & +RCCA~\cite{huang2019ccnet} &   +AlignCM \\
        \toprule[0.2em]
         mIoU(\%)& 77.1 & 79.0  & 79.1 &  79.5 & 79.4 \\
        \bottomrule[0.1em]
    \end{tabular}
    \vspace{-4mm}
    \end{table}

\noindent\textbf{Aligned Context Modeling.}
We also further study the influence of our Aligned Context Modeling module. Results from Aligned Context Modeling with/without alignment are shown in Table~\ref{tab:cityscape_cam},
where ``w/o Alignment'' uses the Aligned Context Modeling without the alignment function, ``w/ Alignment'' uses the Aligned Context Modeling module to augment the feature representation.
Following the settings of PSPNet~\cite{zhao2017pyramid}, we choose ${2,3,6}$ as bin size of the pooled feature map, respectively. The results demonstrate that our method substantially outperforms the baseline in all metrics. 
In addition, we also compare the proposed AlignCM with other context modeling approaches. The results are shown in Table~\ref{tab:ablation_contextual_module}. The baseline is bottom-up aggregation architecture based on ResNet-101. RCCA~\cite{huang2019ccnet} is an attention based context modeling module, which achieves the best performance. The proposed AlignCM achieves 79.4\% mIoU, which is comparable to RCCA~\cite{huang2019ccnet} and outperforms the PSP~\cite{zhao2017pyramid} and ASPP~\cite{chen2018deeplab}.




    \begin{table}[!t]
    \renewcommand{\arraystretch}{1.3}
    \setlength{\tabcolsep}{0.9em}
    \caption{The influence of the DS, OHEM, and MS on the Cityscapes \emph{val}. }
    \label{tab:ablation_other}
    \centering \small
    \vspace{-3mm}
    \begin{tabular}{c c c c}
        \toprule[0.2em]
         DS & OHEM & MS & mIoU (\%)  \\
        \toprule[0.2em]
                  &           &        & 80.5 \\
        $\surd$   &           &        & 80.6 \\
        $\surd$   &  $\surd$  &        & 81.5 \\
        $\surd$   &  $\surd$  & $\surd$  & 82.4 \\
        \bottomrule[0.1em]
    \end{tabular}
    \vspace{-4mm}
    \end{table}

\noindent\textbf{Adding Bells and Whistles.}
To further improve the performance,
we add some standard bells and whistles in our approach,
including Deep Supervision (DS), Online Hard Example Mining (OHEM), and Multi-Scale (MS) testing.
With DS, we employ class-balanced cross-entropy loss on both the final output and the intermediate
feature map $\mathbf{F_3}$. For MS, we apply multiple scales including {0.75x, 1x, 1.25x, 1.5x. 1.75x} with left-right flipping to improve the performance from 81.5\% to 82.4\% on the val set.
All the related results are reported in Table~\ref{tab:ablation_other}.
As we can see, our approach benefits a lot from these standard bells and whistles.

\begin{table}[!t]
        \renewcommand{\arraystretch}{1.3}
        \setlength{\tabcolsep}{0.6em}
        \caption{Result comparison with state of the arts on 
        the Cityscapes \emph{val}.}
        \label{tab:cityscape_val}
        \vspace{-3mm}
        \centering \small
        \begin{threeparttable}
        \begin{tabular}{lccc}
            \toprule[0.2em]
            Approach & Backbone & Multi-scale & mIoU (\%)  \\
            \toprule[0.2em]
            PSPNet~\cite{zhao2017pyramid} & ResNet-101 & No & 78.5  \\
            DeepLabv3~\cite{chen2017rethinking} & ResNet-101 & Yes & 79.3  \\
            DeepLabv3+~\cite{chen2018encoder} & Xception-65 & No & 79.1\\
            DPC~\cite{chen2018searching}~\dag & Xception-71 & No & 80.8\\
            Gated-scnn~\cite{takikawa2019gated} & WiderResNet-38 & No & 80.8 \\
            \midrule
            AlignSeg (Ours) & ResNet-101 & No & 81.5\\
            AlignSeg (Ours) & ResNet-101 & Yes & \textbf{82.4}\\
            \bottomrule[0.1em]
        \end{tabular}
        \begin{tablenotes} 
        \item \dag ~use extra COCO dataset for training.
      \end{tablenotes}
      \end{threeparttable}
      \vspace{-5mm}
    \end{table}

\vspace{-3mm}
\subsubsection{Compare with State-of-the-Arts}

We compare our AlignSeg with the previous approaches on the Cityscapes validation set in Table~\ref{tab:cityscape_val}.
Actually, some of these results should not be compared with our approach directly because these approaches are trained on different (even larger) training sets or use different backbones.
We provide these results for reference.
We can observe that our approach outperforms the approaches using the same ResNet-101 backbone and the same training set~\cite{chen2017rethinking,zhao2017pyramid},
and the approaches using stronger backbone (Xception-65, Xception-71 and WiderResnet-38) and larger training sets (COCO).
Particularly, even without stronger backbones, our approach still achieves better performance than other approaches like DeepLabv3+~\cite{chen2018encoder} and Gated-scnn~\cite{takikawa2019gated},
which confirms the effectiveness of our proposed bottom-up aggregation with the feature alignment method.

\begin{table}[!t]
        \renewcommand{\arraystretch}{1.3}
        \setlength{\tabcolsep}{1.2em}
        \caption{Result comparison with state of the arts on 
        the Cityscapes \emph{test}.}
        \label{tab:cityscape_test}
        \centering \small
        \vspace{-3mm}
        \begin{threeparttable}
        \begin{tabular}{lcc}
            \toprule[0.2em]
            Approach & Backbone & mIoU (\%)  \\
            \toprule[0.2em]
            \multicolumn{3}{l}{\textit{dilated convlution based approaches}} \\
            DeepLab-v2~\cite{chen2018deeplab}~\dag  & ResNet-101 & 70.4  \\
            ResNet-38~\cite{wu2016wider}~\dag & WiderResnet-38 & 78.4\\
            PSPNet~\cite{zhao2017pyramid}~\dag & ResNet-101 & 78.4 \\
            AAF~\cite{ke2018adaptive}~\dag & ResNet-101 & 79.1 \\
            DUC~\cite{wang2018understanding}~\ddag & ResNet-101 & 77.6\\
            SAC~\cite{zhang2017scale}~\ddag & ResNet-101 & 78.1\\
            PSANet~\cite{zhao2018psanet}~\ddag & ResNet-101 & 80.1 \\
            DenseASPP~\cite{yang2018denseaspp}~\ddag & DenseNet-161 & 80.6\\
            CCNet~\cite{huang2019ccnet}~\ddag & ResNet-101 & 81.4 \\
            DANet~\cite{fu2019dual}~\ddag & ResNet-101 & 81.5 \\
            \midrule
            \multicolumn{3}{l}{\textit{encoder-decoder based approaches}} \\
            RefineNet~\cite{lin2017refinenet}~\ddag & ResNet-101 & 73.6\\
            GCN~\cite{peng2017large}~\ddag  & ResNet-101 & 76.9\\
            BiSeNet~\cite{yu2018bisenet}~\ddag & ResNet-101 & 78.9 \\
            DFN~\cite{yu2018learning}~\ddag & ResNet-101 & 79.3 \\
            SPGNet~\cite{cheng2019spgnet}~\ddag & ResNet-101 & 81.1 \\
            ACNet~\cite{fu2019adaptive}~\ddag & ResNet-101 & 82.4 \\
            \midrule
            \multicolumn{3}{l}{\textit{bottom-up aggregation based approaches}} \\
            FRRN~\cite{pohlen2017full}~\dag & FRRN-B & 71.8 \\
            DLA~\cite{yu2018deep}~\dag & DLA-169 & 75.9\\
            SeENet~\cite{pang2019towards}~\ddag & ResNet-101 & 81.2\\
            AlignSeg (Ours)~\dag  & ResNet-101 & \textbf{81.5} \\
            AlignSeg (Ours)~\ddag & ResNet-101 & \textbf{82.6}\\
            \bottomrule[0.1em]
        \end{tabular}
        \begin{tablenotes}
        \item \dag ~only use the train-fine for training.
        \item \ddag ~use both the train-fine and val-fine for training.
        \end{tablenotes}
      \end{threeparttable}
      \vspace{-5mm}
    \end{table}


We further compare our approach with other state-of-the-arts on the Cityscapes test set in Table~\ref{tab:cityscape_test}.
It is obvious that our approach achieves better performance than all previous approaches based on ResNet-101 or stronger backbones.
Our approach without using the val set for training obtains even better performance than other approaches that use the val set for training.
Especially, the state-of-the-art dilated convolution based method DenseASPP~\cite{yang2018denseaspp}, which even uses a stronger backbone DenseNet-161~\cite{huang2017densely}, performs worse than our approach.
The state-of-the-art encoder-decoder based method ACNet~\cite{fu2019adaptive}, which introduces an adaptive way to fuse the global and local context, also does not perform as well as our approach.
In addition, our approach obtains much better performance than previous bottom-up aggregation based methods~\cite{pohlen2017full,yu2018deep,pang2019towards},
due to our designs including the Aligned Feature Aggregation module and Aligned Context Modeling module.
    
\vspace{-3mm}
\subsection{Experiments on ADE20K}
Here we show experiments on the large-scale ADE20K dataset.
Since the annotations for the testing set have not been released,
we only compare our approach with the previous state-of-the-arts on the ADE20K validation (val) set
in Table~\ref{tab:ade20k}.
As we can see, our approach outperforms all previous ResNet-101 based approaches~\cite{liang2018dynamic,xiao2018unified,zhang2018context,zhang2017scale,zhao2017pyramid,zhao2018psanet,fu2019adaptive} and even the ResNet-152 based approach~\cite{lin2017refinenet}.
In particular, our approach obtains 45.95\% mIoU, which surpasses the previous best-performed approach ACNet~\cite{fu2019adaptive}.


    \begin{table}[!t]
        \renewcommand{\arraystretch}{1.3}
        \setlength{\tabcolsep}{1.8em}
        \caption{Result comparison with state of the arts on the ADE20K \emph{val}.}
        \label{tab:ade20k}
        \vspace{-3mm}
        \centering \small
        \begin{tabular}{lcc}
            \toprule[0.2em]
            Approach & Backbone & mIoU (\%)  \\
            \toprule[0.2em]
            RefineNet~\cite{lin2017refinenet} & ResNet-152 & 40.70\\
            UperNet~\cite{xiao2018unified} & ResNet-101 & 42.66 \\
            PSPNet~\cite{zhao2017pyramid} & ResNet-101 & 43.29 \\
            DSSPN~\cite{liang2018dynamic} & ResNet-101 & 43.68 \\
            PSANet~\cite{zhao2018psanet} & ResNet-101 & 43.77 \\
            SAC~\cite{zhang2017scale} & ResNet-101 & 44.30\\
            EncNet~\cite{zhang2018context} & ResNet-101 & 44.65 \\
            CCNet~\cite{huang2019ccnet} & ResNet-101 & 45.22 \\
            ACNet~\cite{fu2019adaptive} & ResNet-101 & 45.90 \\
            \midrule
            AlignSeg (Ours) & ResNet-101 & \textbf{45.95}\\
            \bottomrule[0.1em]
        \end{tabular}
        \vspace{-2mm}
    \end{table}

\vspace{-3mm}
\subsection{Experiments on COCO}
To further demonstrate the generalization ability of AlignSeg, we conduct the instance segmentation task on COCO~\cite{lin2014microsoft} using the Mask R-CNN~\cite{he2017mask} as the baseline. We modify the Mask R-CNN backbone by adding the AlignFA module in each feature aggregation, denoted as ``R-101-FPN-AlignFA''. We evaluate a standard baseline of ResNet-101. All models are fine-tuned from ImageNet pre-training. We report the results in terms of box AP and mask AP in Table~\ref{tab:coco}. The results show that our method outperforms the baseline in all metrics, particularly in mask AP. 

  \begin{table}[!t]
    \renewcommand{\arraystretch}{1.3}
    \setlength{\tabcolsep}{0.8em}
    \centering
    \caption{Comparisons on COCO \emph{val}.}
    \label{tab:coco}
    \vspace{-3mm}
    \centering \small
    \begin{tabular}{l c c}
        \toprule[0.2em]
         Backbone & box AP & mask AP \\
        \toprule[0.2em]
         R-101-FPN & 39.4 &    35.9 \\
         R-101-FPN-AlignFA & 40.0 &    37.2 \\
        \bottomrule[0.1em]
    \end{tabular}
    \vspace{-4mm}
    \end{table}


\begin{figure*}[!t]
    \centering
    \includegraphics[width=0.85\linewidth]{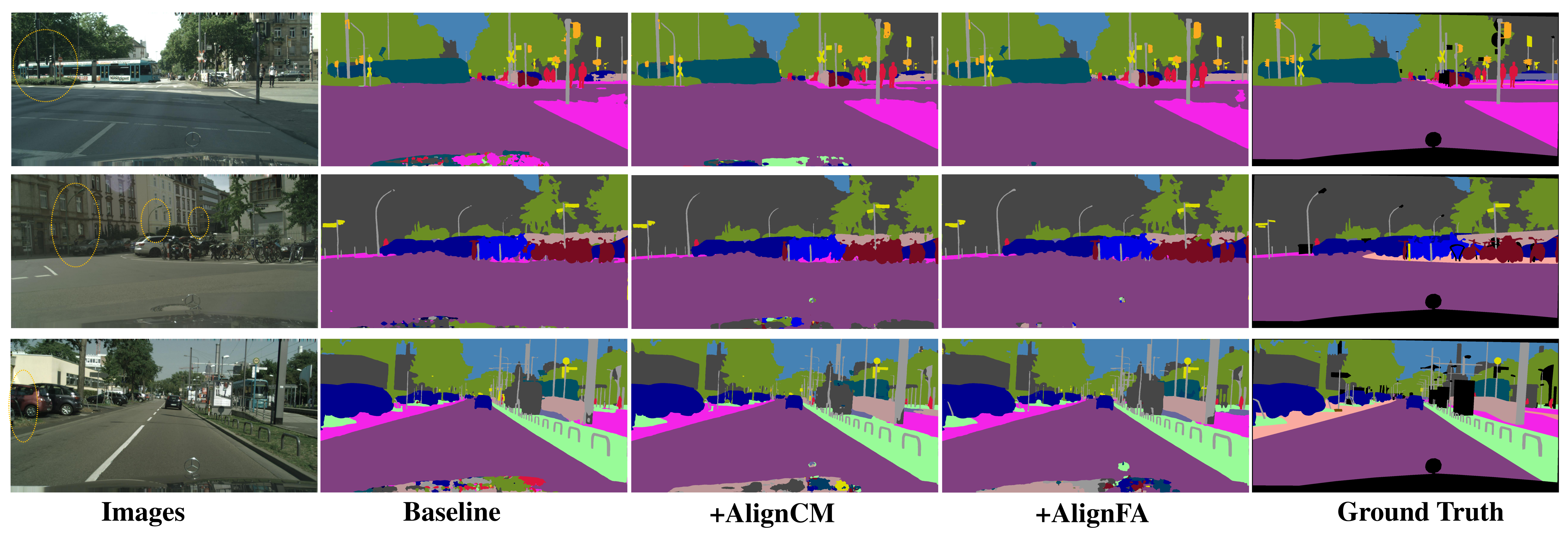}
    \vspace{-5mm}
    \caption{Some visualization comparisons among different approaches on the Cityscapes \emph{val} set.
    The first to the fifth columns are original images, results from the baseline, results from baseline with AlignCM,  results from the baseline with AlignCM and AlignFA, and ground truth results, respectively.
    The regions inside the yellow dashed circles are some boundary regions.  Best to zoom in.}
    \label{fig:vis}
    \vspace{-6mm}
\end{figure*}

\begin{figure}[!t]
        \centering
        \includegraphics[width=0.7\linewidth]{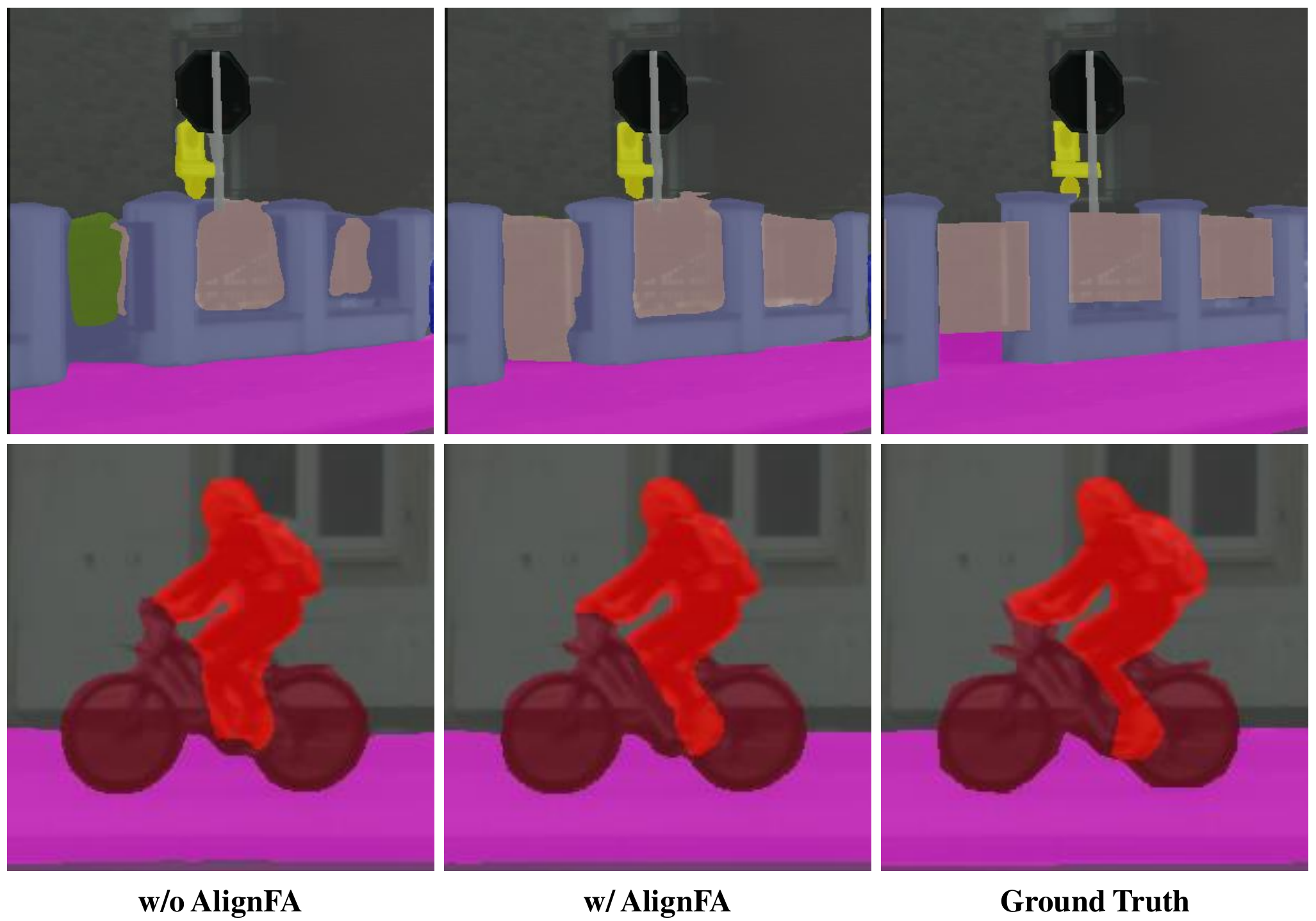}
        \vspace{-2mm}
        \caption{Some visualization comparisons between our approach w/o and w/ Aligned Feature Aggregation (AlignFA) of some cropped patches on the Cityscapes \emph{val}.}
        \label{fig:faa_vis}
        \vspace{-6mm}
    \end{figure}

\vspace{-3mm}
\subsection{Visualization}
We present some visualization comparisons among the bottom-up aggregation architecture (baseline), the bottom-up aggregation architecture with AlignCM, and the bottom-up aggregation architecture with AlignCM and AlignFA on the Cityscapes validation (val) set in Fig.~\ref{fig:vis}.
As we can see, our bottom-up aggregation architecture with AlignCM produces higher-quality segmentation results than the bottom-up aggregation architecture and reduce unexpected misclassification, \ie, the first row in Fig.~\ref{fig:vis}, the bus inside the yellow dashed circles, which is misclassified into car by baseline model, is classified correctly with the help of AlignCM.
After adding the Aligned Feature Aggregation module, our approach produces finer segmentation results than the approaches without AlignFA, especially for the boundary regions (the regions inside the yellow dashed circles). We also randomly crop some patches from some images in the Cityscapes val set,
and show the segmentation results w/o and w/ AlignFA in Fig.~\ref{fig:faa_vis}.
These visualization results further demonstrate that our AlignFA can provide more accurate predictions for boundary regions.

More visualization can be found in the supplementary materials. Especially, it is interesting to see that the offset maps ($\Delta^F$ and $\Delta^A$) are significantly different. They have the different direction and intensity of the alignment at the same point, indicating that the predicted offset maps attempt to correct the misalignment between feature $F$ and feature $A$ and deform both feature maps into a more canonical view.


\vspace{-3.5mm}
\section{Conclusion and Future Work} \label{Conclusion}

In this paper, we focus on the feature misalignment issue
in previous popular feature aggregation architectures for semantic segmentation, thus we propose AlignSeg, which has two primary components:  Aligned Feature Aggregation (AlignFA) and Aligned Context Modeling (AlignCM). AlignFA is used to precisely align both high-resolution and low-resolution feature maps for multi-resolution feature aggregation.  AlignCM collects adaptive, selective, and rich contextual information by integrating the learnable alignment approach with pooling based methods.
Experimental results show that our approach obtains state-of-the-art performance on two challenging semantic segmentation datasets
(82.6\% and 45.95\% mIoU scores on the Cityscapes test set and ADE20K validation set respectively).
We believe that our approach should also be able to handle other pixel-level prediction tasks such as edge or saliency detection, which we would like to explore in the future. 

\vspace{-3.5mm}
\section*{Acknowledgement} \label{Conclusion}
This work was in part supported by NSFC (No. 61876212 and No. 61733007), China Scholarship Council, and IBM-ILLINOIS Center for Cognitive Computing Systems Research (C3SR) - a research collaboration as part of the IBM AI Horizons Network.

\vspace{-3.5mm}


\ifCLASSOPTIONcaptionsoff
  \newpage
\fi



%
{\small
    \bibliographystyle{IEEEtran}
    \bibliography{AlignSeg}
}

\end{document}